\documentclass[Preprint,12pt]{elsarticle}
\usepackage{newtxtext}

\usepackage{lineno}

\journal{Expert Systems with Applications}

\usepackage[bottom]{footmisc} % 添加此行以将脚注固定在底部

\usepackage{geometry}
\usepackage{amsmath}
\usepackage{algorithm}
\usepackage{algpseudocode}
\usepackage{url}
\usepackage{ upgreek }
\usepackage{pdflscape}
\usepackage{caption}
\usepackage{setspace}
\usepackage{pifont}
\usepackage{amsmath}
\usepackage{bbding}
\usepackage{amsfonts} 
\usepackage{amssymb}
\usepackage{mathrsfs}

\usepackage{color}
\usepackage{ dsfont }
\usepackage[misc]{ifsym}

\usepackage{amsmath}      % 加载amsmath宏包
\usepackage{graphicx}  
\usepackage{array}
\usepackage{verbatim}
\usepackage{titletoc}
\usepackage{color,xcolor}

\usepackage{units}
\usepackage{multirow}
\usepackage{tabularx}
\usepackage{booktabs}
\usepackage{bbm}
\usepackage{amsmath}      % 加载amsmath宏包
\usepackage{bm} 
\usepackage{titlesec}
\usepackage{algorithm}
\usepackage{algpseudocode}
\usepackage{fvextra}
\titleformat{\paragraph}
  {\normalfont\normalsize\itshape}{\theparagraph}{1em}{}
\titlespacing*{\paragraph}
  {0pt}{1.5ex plus 0.5ex minus .2ex}{0.5ex plus .2ex}  % 调整标题与下一段之间的间距

\definecolor{mediumgray}{gray}{0.88}
\definecolor{lightgray}{gray}{0.95}

\usepackage{colortbl}

\usepackage{amsmath}

\usepackage{xcolor}
\definecolor{DeBlue}{RGB}{0,0,0}
\definecolor{DeGreen}{RGB}{84,130,53}
\definecolor{DeBlack}{RGB}{0,0,0}

\usepackage{natbib}
\usepackage{fancyhdr}

\begin{document}

\title{Solar-VLM: Multimodal Vision-Language Models for Augmented Solar Power Forecasting} 

\begin{frontmatter}

\author[inst1]{Hang Fan}
% fanhang@ncepu.edu.cn
\affiliation[inst1]{organization={School of Economic and Management},
            addressline={North China Electric Power University}, 
            city={102206, Beijing},
            country={China}}

\author[inst2]{Haoran Pei}
% haoranpei@ncepu.edu.cn
\affiliation[inst2]{organization={School of Control and Computer Engineering},
            addressline={North China Electric Power University}, 
            city={102206, Beijing},
            country={China}}

\author[inst3]{Runze Liang}
% liangrz23@mails.tsinghua.edu.cn
\affiliation[inst3]{organization={Department of Electrical Engineering},
             addressline={Tsinghua University},
             city={100084, Beijing},
            country={China}}

\author[inst4]{Weican Liu}
% weican001@e.ntu.edu.sg
\affiliation[inst4]{organization={School of Electrical and Electronic Engineering},
             addressline={Nanyang Technological University}, 
             city={50 Nanyang Avenue, 639798},
             country={Singapore}}

\author[inst2]{Long Cheng\corref{cor1}}
% lcheng@ncepu.edu.cn
\cortext[cor1]{Corresponding author.}

\author[inst3]{Wei Wei\corref{cor2}}
% wei-wei04@mails.tsinghua.edu.cn
\cortext[cor2]{Principal corresponding author.}

\normalsize

\begin{abstract}
Photovoltaic (PV) power forecasting plays a critical role in power system dispatch and market participation. Because PV generation is highly sensitive to weather conditions and cloud motion, accurate forecasting requires effective modeling of complex spatiotemporal dependencies across multiple information sources. Although recent studies have advanced AI-based forecasting methods, most fail to fuse temporal observations, satellite imagery, and textual weather information in a unified framework. This paper proposes Solar-VLM, a large-language-model-driven framework for multimodal PV power forecasting. First, modality-specific encoders are developed to extract complementary features from heterogeneous inputs. The time-series encoder adopts a patch-based design to capture temporal patterns from multivariate observations at each site. The visual encoder, built upon a Qwen-based vision backbone, extracts cloud-cover information from satellite images. The text encoder distills historical weather characteristics from textual descriptions. Second, to capture spatial dependencies across geographically distributed PV stations, a cross-site feature fusion mechanism is introduced. Specifically, a Graph Learner models inter-station correlations through a graph attention network constructed over a K-nearest-neighbor (KNN) graph, while a cross-site attention module further facilitates adaptive information exchange among sites. Finally, experiments conducted on data from eight PV stations in a northern province of China demonstrate the effectiveness of the proposed framework. Our proposed model is publicly available at \url{https://github.com/rhp413/Solar-VLM}.
\end{abstract}


\end{frontmatter}
\setcounter{page}{1}

\section{Introduction}
% \linenumbers

\subsection{Background and motivation}

Driven by growing concerns over excessive consumption of fossil fuels and environmental pollution, renewable energy sources have attracted worldwide attention. Among them, photovoltaic (PV) energy has become one of the most widely deployed technologies for renewable power generation and grid integration \cite{pv_widely_usage}. However, the inherent intermittency and uncertainty of solar energy introduce significant challenges to maintaining power system balance and operational reliability. As a result, accurate PV power forecasting has become increasingly important, serving as one of the most economical and practical means of supporting large-scale PV integration into the power grid \cite{pv_forecast_important}. Over the past decades, extensive efforts have been devoted to PV power forecasting, and the existing methods can generally be classified into two categories: traditional approaches and deep learning-based approaches.

\subsection{Traditional Prediction Methods}

Traditional forecasting approaches can be broadly categorized into physical methods and statistical methods. Physical forecasting methods rely on meteorological inputs, such as temperature and rainfall, and generate predictions based on established physical models \cite{physics_regional_pv,Inman2013Solar}. Although these methods offer strong physical interpretability, their performance largely depends on the accuracy of weather forecasts and the system parameterization \cite{sota_solar_review}.

Statistical models, such as autoregressive integrated moving average \cite{arima_pv_prediction}, support vector machine \cite{SVM_pv_prediction}, random forest \cite{random_forest_pv}, and gradient boosted regression trees \cite{gbrt_pv}, instead learn empirical relationships directly from historical data. These methods are generally easy to implement and computationally efficient. However, they often have limited ability to capture the complex nonlinear dependencies inherent in PV power forecasting, including both the temporal dynamics of PV output and the interactions between meteorological variables and PV generation.

\subsection{Deep Learning-based Prediction Methods}

In recent years, deep learning has significantly advanced PV power forecasting by enabling end-to-end modeling of complex nonlinear relationships from high-dimensional data. According to the input modalities and forecasting objectives, existing studies can generally be grouped into three main categories: (1) time-series-based forecasting, (2) spatiotemporal forecasting across multiple PV sites, and (3) vision-driven and multimodal forecasting.

\subsubsection{Time Series–based PV Forecasting}

The most direct deep learning paradigm for PV power forecasting models the mapping from historical time-series inputs, such as past PV output, global irradiance, and temperature, to future PV power generation. In \cite{rnn_pv_prediction}, several deep neural architectures for time-series-based PV forecasting were developed and systematically compared, including Long Short-Term Memory (LSTM), Bidirectional LSTM (BiLSTM), Gated Recurrent Unit (GRU), Bidirectional GRU (BiGRU), and one-dimensional convolutional neural network (1D-CNN). Li et al. \cite{wavelet_pv_prediction} proposed a hybrid framework that combines wavelet packet decomposition with LSTM networks for one-hour-ahead forecasting at five-minute resolution. Tao et al. \cite{transformer_pv_prediction} further developed a transformer-based PV forecasting model that integrates two types of temporal information, namely historical local measurements and future numerical weather prediction (NWP) data. Despite their effectiveness, these methods are limited to single-site temporal modeling and do not exploit broader spatial or multimodal information, which may constrain their forecasting performance.

\subsubsection{Spatio-Temporal Forecasting across Multiple PV Sites}

The outputs of spatially distributed PV plants exhibit strong correlations driven by shared weather patterns and cloud motion. Capturing these spatio-temporal dependencies is crucial for accurate forecasting. Graph neural networks (GNNs), with their ability to explicitly model relational structures among multiple sites, are particularly well suited for this task.
In \cite{DEST_GNN}, an adaptive graph convolutional network (GCN) was designed to capture the spatio-temporal correlations of multiple PV sites. The model employs an adaptive adjacency matrix and a temporal convolutional component to achieve accurate intra-hour PV power forecasting. Zhang et al. \cite{optimal_graph_structure} clustered PV power plants with similar solar resource conditions to improve intra-class compactness and inter-class separability, and further proposed a novel graph connectivity index to guide graph construction, thereby enhancing the effectiveness of GNN–based forecasting models. An explainable energy forecasting framework based on spatio-temporal graph neural networks was proposed in \cite{Verdone2024Explainable}. In this framework, temporal and spatial information are processed through a one-dimensional CNN and a spectral graph convolution, respectively, to generate forecasts. Furthermore, a GNN explainer is applied to the forecasting model to provide insights into the generation process and improve interpretability.

\subsubsection{Vision-driven and Multimodal Forecasting}

Integrating visual and multimodal information into PV power forecasting provides richer contextual understanding of PV generation patterns, complementing purely numerical or time-series data. Consequently, this direction has attracted increasing research interest for its potential to capture complex atmospheric dynamics and improve forecasting accuracy.
Fu et al. \cite{cae_pv_predict} proposed convolutional autoencoder–based sky image prediction models to overcome the deficiencies of traditional digital image processing technology–based sky image prediction models, including limited input image sequence length and linear image extrapolation. Xu et al. \cite{ViT_PV_prediction} utilized ground-based whole-sky cameras to monitor clouds above PV farms and develop a cloud image-based ultra-short-term forecasting framework. Within this framework, an integration of the Vision Transformer (ViT) model and the GRU encoder is designed for high-dimensional latent feature analysis. Qin et al. \cite{intergrate_ground_and_satellite} developed a deep learning–based hybrid framework that leverages both ground and satellite observations to improve PV power forecasting. Specifically, CNNs are employed to extract cloud motion information from satellite imagery, and LSTM networks are subsequently used to capture the long-range relationships between cloud dynamics and PV power generation. To address the data imbalance problem in CNN-based PV power prediction, where the dataset typically contains abundant sunny-condition images but insufficient cloudy-condition images, Nie et al. \cite{data_augmentation_pv} proposed a method for the enrichment and augmentation of an imbalanced sky image dataset. 

With the rapid development of large language models (LLMs) like GPT \cite{GPT} and LLaMA \cite{LLaMA}, leveraging their learned textual knowledge to assist time series forecasting has become an emerging research direction. 
Some studies \cite{LLMTIME, llm4ts, one_fits_all} attempted to directly tokenize or embed time series data for input into LLMs to perform forecasting tasks. Although this paradigm is relatively straightforward, its input representation deviates from natural language forms, which constrains the model’s ability to fully exploit the pretrained linguistic knowledge. To overcome this limitation, recent approaches \cite{TimeLLM, UniTime} incorporate domain knowledge, task descriptions, and time-series statistics into natural language prompts, allowing LLMs to better utilize their pretrained semantic understanding for forecasting.

Beyond purely text-based architectures, vision–language models (VLMs) such as ViLT \cite{ViLT}, Qwen \cite{CLIP}, and LLaVA \cite{LLaVA} jointly learn from textual and visual modalities through multimodal pretraining, thereby bringing greater potential for pretrained models in PV power forecasting. Building upon this paradigm, Lin et al. \cite{pv_vlm} developed an integrated PV forecasting framework that jointly utilizes visual, textual, and time-series information to capture cross-modal dependencies. In their framework, a pretrained VLM is employed to extract semantic and spatial representations from sky images, while an LLM encodes textual prompts generated from historical statistics and dataset descriptions.

\subsection{Contributions}

Although substantial progress has been achieved in time-series-based PV forecasting, spatiotemporal correlation modeling, and multimodal prediction, these advances have largely evolved in isolation. A unified framework that jointly exploits temporal dynamics, inter-site spatial dependencies, and heterogeneous multimodal information remains unexplored in the existing literature. To fill this gap, this paper proposes \textbf{Solar-VLM}, a novel vision-language-time-series framework for PV power forecasting. The main contributions of this work are summarized as follows:

\begin{itemize}

\item \textbf{A unified multi-site multimodal forecasting paradigm.} To overcome the limitations of conventional PV forecasting methods that rely on a single modality, we propose \textbf{Solar-VLM}, a unified framework that jointly integrates satellite imagery (visual modality), historical statistical text (language modality), and temporal generation data (time-series modality) for multi-site PV forecasting. By deeply fusing these heterogeneous information sources, the proposed framework enables collaborative learning across both modalities and sites, thus establishing a new paradigm for multimodal spatiotemporal PV forecasting.

\item \textbf{Cross-modal interaction and spatiotemporal dependency modeling.}  
To learn informative multimodal representations and capture complex inter-site correlations, Solar-VLM leverages the pretrained Qwen model to extract semantic features from visual and textual inputs, benefiting from prior knowledge embedded in pretrained foundation models. Building upon this, a cross-modal attention mechanism and a graph learner module have been developed. The former enables fine-grained interaction and dependency modeling across modalities, while the latter explicitly characterizes the relational structure among PV sites. Together, these components allow Solar-VLM to capture intricate nonlinear dependencies that are difficult to represent using conventional approaches.

\item \textbf{Retrieval-augmented long-term historical information utilization.}  To mitigate the loss of long-range historical information commonly encountered in deep learning-based forecasting, a retrieval-augmented learning module with a dual-memory architecture is designed, consisting of local memory and global memory. In addition, a memory fusion gate is introduced to dynamically retrieve and integrate critical historical information from both memory spaces. This design substantially strengthens the model’s ability to exploit long-term dependencies and further improves forecasting accuracy.
\end{itemize}

\section{Problem Formulation}

We consider a multisite, multimodal PV power forecasting problem involving a set of $M$ PV plants. For each site $i$, multiple physical variables are observed at each time step $t$, such as global horizontal irradiance, temperature, and wind speed. These measurements, together with the corresponding PV power output, are represented by a feature vector $\mathbf{x}_t^{(i)} \in \mathbb{R}^{D}$, where $D$ denotes the number of numerical features. The historical numerical observations used for forecasting are denoted as $\mathbf{X}_{1:L} = \{\mathbf{x}_t^{(i)} \mid i = 1,\dots,M;\; t = 1,\dots,L\}$, where $L$ denotes the length of the historical observation window.

In addition to numerical measurements, satellite imagery capturing atmospheric conditions around each PV site is incorporated into the forecasting process. Let $\mathbf{S}_{L-k+1:L} = \{\mathbf{s}_t^{(i)} \mid i = 1,\dots,M;\; t = L-k+1,\dots,L\}$ denote the sequence of the most recent $k$ satellite images prior to the prediction horizon. Each image  $\mathbf{s}_t^{(i)}$ focuses on the local region surrounding a specific PV plant and provides visual information about cloud cover and atmospheric dynamics relevant to PV generation.

Furthermore, a textual description is available for each site at the current time step $L$, describing the local meteorological conditions. We denote the collection of textual inputs as $\mathbf{C}_L = \{\mathbf{c}_L^{(i)} \mid i = 1,\dots,M\}$, which provides high-level semantic context for the forecasting task.

The objective is to predict the future PV power outputs at all sites over the next $T$ time steps, given the historical numerical observations, satellite imagery, and textual descriptions. Formally, the forecasting problem is defined as
\begin{equation*}
\mathbf{p}_{L+1:L+T}
=
f\!\left(
\mathbf{X}_{1:L},
\mathbf{S}_{L-k+1:L},
\mathbf{C}_L
\right),
\end{equation*}
where $\mathbf{p}_{L+1:L+T} = \{\mathbf{p}_{L+1:L+T}^{(i)}\}_{i=1}^M$ denotes the predicted PV power output sequences for all sites, and $f(\cdot)$ represents the proposed forecasting model.

\section{Methodology}

\begin{figure}[t!]
    \centering
    \includegraphics[width=1.0\linewidth]{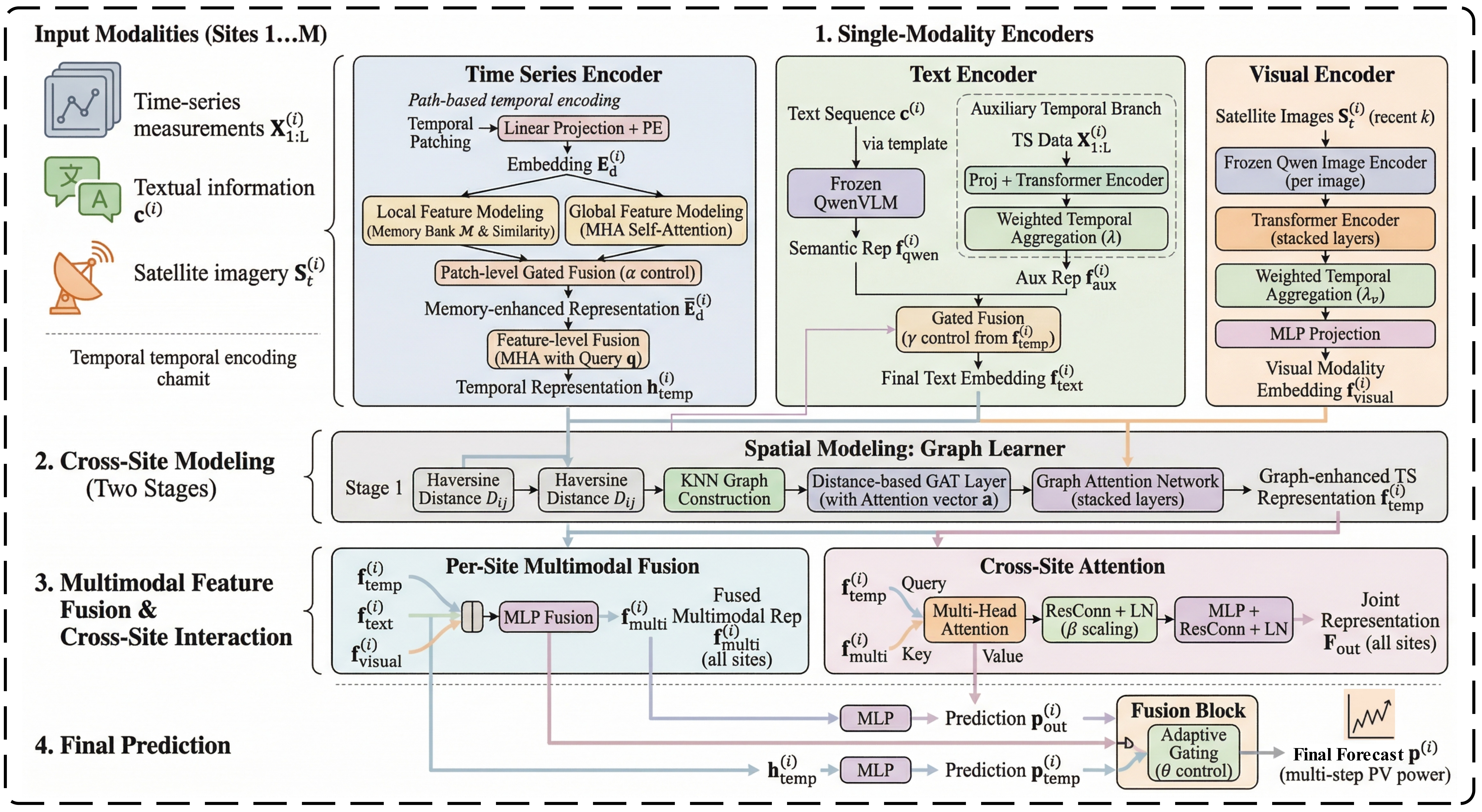}
    \caption{Overview of the proposed model.}
    \label{fig:framework}
\end{figure}

We propose a unified multimodal and multi-site forecasting framework for PV power prediction, as illustrated in Fig.~\ref{fig:framework}. For each PV site, the time-series measurements, satellite imagery, and textual information are first encoded independently using modality-specific encoders, yielding latent representations that capture the intrinsic characteristics of each modality. Subsequently, a per-site multimodal fusion module integrates the encoded representations from the three modalities to produce a unified multimodal representation for each site.

To support cross-site modeling, we introduce two complementary mechanisms. First, a graph learner is applied to the encoded time-series features prior to multimodal fusion to infer structured inter-site relationships, with the design rationale discussed in Subsection~\ref{sec:cross_site_modeling}. Second, after per-site multimodal fusion, a cross-site attention module models spatial correlations among sites based on the fused multimodal representations, enabling flexible information sharing across sites. The resulting multimodal predictions are further fused with those directly obtained from the time-series encoder to generate the final multi-step photovoltaic power forecasts.

\subsection{Single-Modality Encoders}
\subsubsection{Time Series Encoder}

The time series encoder aims to transform the original numerical observations into informative latent representations.
In PV power forecasting, time series data at each site typically exhibit significant local temporal patterns.
To effectively model such structures, we adopt a patch-based modeling strategy. First, the time series of each variable is independently partitioned into temporal patches and encoded as a patch-level sequence. Second, a dual-path modeling mechanism, including local feature modeling and global feature modeling, is applied to each patch sequence to capture local temporal patterns within patches and long-range temporal dependencies across patches.
The two paths are then fused at the patch level using a gated mechanism.
Finally, feature-level fusion with multi-head attention is performed to aggregate information across variables, yielding the final temporal representation for each site.

\textbf{Patch-based Temporal Encoding.} For each site $i$, we reorganize the historical numerical observations
$\mathbf{X}^{(i)}_{1:L}$ into a feature-wise temporal representation
$\mathbf{Z}^{(i)} = \big( \mathbf{z}^{(i)}_1, \ldots, \mathbf{z}^{(i)}_D \big)$,
where each vector $\mathbf{z}^{(i)}_d \in \mathbb{R}^{L}$ corresponds to the temporal sequence of the $d$-th variable at site $i$.
Each feature-wise sequence is independently partitioned into overlapping temporal patches with patch length $l_p$ and stride $s_p$, such that $\mathbf{z}^{(i)}_d$ is transformed into
$N_p = \left\lfloor \frac{L - l_p}{s_p} \right\rfloor + 1$ patches, denoted by
$\mathbf{P}^{(i)}_d \in \mathbb{R}^{N_p \times l_p}$.
In each patch, a linear projection is applied, and a corresponding positional encoding is added, yielding the patch-level sequence embedding
$\mathbf{E}^{(i)}_d \in \mathbb{R}^{N_p \times d_{\mathrm{model}}}$:
\begin{equation*}
\mathbf{E}^{(i)}_d = \mathrm{MLP}(\mathbf{P}^{(i)}_d) + \mathbf{PE},
\end{equation*}
where $d_{\mathrm{model}}$ denotes the embedding dimension, and $\mathbf{PE}$ is the sinusoidal positional encoding defined as
\begin{align*}
\mathbf{PE}_{(p,\,2k)} &= \sin\!\left( \frac{p}{10000^{2k / d_{\mathrm{model}}}} \right), \\
\mathbf{PE}_{(p,\,2k+1)} &= \cos\!\left( \frac{p}{10000^{2k / d_{\mathrm{model}}}} \right),
\end{align*}
where $p$ denotes the patch index and $k$ is the dimension index.

\textbf{Local Feature Modeling.} This module aims to enhance patch-level representations by retrieving relevant historical patterns from a shared memory bank.
This design allows each temporal patch to explicitly leverage previously observed dynamics beyond the current input window.
To this end, a fixed-size memory
$\mathcal{M}\in\mathbb{R}^{N_m\times d_{\mathrm{model}}}$
is maintained to store historical patch embeddings.
At each step, newly generated patch embeddings are enqueued into the memory bank while the oldest entries are dequeued, following a first-in-first-out circular update strategy.

For each variable $d$, local memory interaction is performed independently at the patch level.
Let $\mathbf{E}_{d,p}^{(i)} \in \mathbb{R}^{d_{\mathrm{model}}}$ denote the $p$-th patch embedding in $\mathbf{E}_d^{(i)}$, and cosine similarity is used to measure the relevance between $\mathbf{E}_{d,p}^{(i)}$ and each memory entry $\mathbf{m}_j \ (j=1,\ldots,N_m)$.
The $k$ memory entries with the highest similarity are retrieved and denoted by
$\mathcal{R}_{d,p}^{(i)} \subset \mathcal{M}$.
The retrieved entries are then transformed by a two-layer MLP and averaged to form a patch-specific local memory context
\begin{equation*}
\mathbf{c}_{d,p}^{(i)} = \frac{1}{k} \sum_{\mathbf{m}_j \in \mathcal{R}_{d,p}^{(i)}} \mathrm{MLP}(\mathbf{m}_j).
\end{equation*}

Finally, the local-enhanced patch representation is obtained via a residual connection between the original patch embedding and its corresponding memory context:
\begin{equation*}
\hat{\mathbf{E}}_{d,p}^{(i)} = \mathbf{E}_{d,p}^{(i)} + \mathbf{c}_{d,p}^{(i)} .
\end{equation*}
Applying this process to all patches yields a sequence of memory-enhanced patch representations $\hat{\mathbf{E}}_{d}^{(i)} \in \mathbb{R}^{N_p \times d_{\mathrm{model}}}$.

\textbf{Global Feature Modeling.}
This module aims to capture long-range temporal dependencies by applying standard multi-head self-attention independently to each $\mathbf{E}_d^{(i)}$:
\begin{equation*}
\tilde{\mathbf{E}}_{d}^{(i)} = \mathrm{MultiHeadAttn}(\mathbf{E}_d^{(i)},\mathbf{E}_d^{(i)},\mathbf{E}_d^{(i)}),
\end{equation*}
where $\mathrm{MultiHeadAttn}(\cdot)$ denotes the standard multi-head self-attention mechanism.
Specifically, given query, key, and value matrices $\mathbf{Q}$, $\mathbf{K}$, and $\mathbf{V}$, multi-head attention is defined as
\begin{equation*}
\mathrm{MultiHeadAttn}(\mathbf{Q}, \mathbf{K}, \mathbf{V})
= [\mathrm{head}_1;\,\ldots;\,\mathrm{head}_h]\mathbf{W}^O,
\end{equation*}
where $[\cdot;\cdot]$ denotes the concatenation operation.
Each attention head is computed as
\begin{equation*}
\mathrm{head}_j = \mathrm{Attention}\big(\mathbf{Q}\mathbf{W}_j^Q,\,
\mathbf{K}\mathbf{W}_j^K,\,
\mathbf{V}\mathbf{W}_j^V\big),
\end{equation*}
and the scaled dot-product attention is given by
\begin{equation*}
\mathrm{Attention}(\mathbf{Q}, \mathbf{K}, \mathbf{V})
= \mathrm{softmax}\!\left(\frac{\mathbf{Q}\mathbf{K}^\top}{\sqrt{d_k}}\right)\mathbf{V}.
\end{equation*}

\textbf{Patch-level Fusion.} After local and global feature modeling, a gated fusion mechanism is adopted to adaptively combine the two representations at the patch level.
For each patch, a MLP takes the corresponding local and global features as input and produces fusion weights $\alpha_{d,p}^{(i)}\in[0,1]$ that control their relative contributions
\begin{equation*}
\alpha_{d,p}^{(i)}=\sigma\left(\mathrm{MLP}([\hat{\mathbf{E}}_{d,p}^{(i)};\tilde{\mathbf{E}}_{d,p}^{(i)}])\right)
\end{equation*}
where $\sigma(\cdot)$ denotes the sigmoid function.

The final memory-enhanced representation is computed as a weighted combination of local-enhanced features and global contextual features:
\begin{equation*}
\bar{\mathbf{E}}_{d,p}^{(i)}
= \alpha_{d,p}^{(i)} \, \hat{\mathbf{E}}_{d,p}^{(i)}
+ \big(1-\alpha_{d,p}^{(i)}\big) \, \tilde{\mathbf{E}}_{d,p}^{(i)},
\end{equation*}

\textbf{Feature-level Fusion.}
After patch-level gated fusion, the fused patch representations
$\bar{\mathbf{E}}_{d,p}^{(i)}$
are stacked to form $\bar{\mathbf{E}}^{(i)} \in \mathbb{R}^{D \times N_p \times d_{\mathrm{model}}}$.
The patch representations of each variable are aggregated by an MLP, which maps the corresponding
$N_p \times d_{\mathrm{model}}$
features to a single
$d_{\mathrm{model}}$-dimensional vector.
This yields a variable-level representation matrix
$
\mathbf{J}^{(i)} \in \mathbb{R}^{D \times d_{\mathrm{model}}},
$
where each row corresponds to one variable.

To model inter-variable dependencies and perform feature-level fusion, 
$\mathbf{J}^{(i)}$ is treated as a sequence and attended by a learnable query vector 
$\mathbf{q} \in \mathbb{R}^{d_{\mathrm{model}}}$ via multi-head attention. 
The resulting temporal representation for site $i$, denoted by 
$\mathbf{h}_{\mathrm{temp}}^{(i)} \in \mathbb{R}^{d_{\mathrm{model}}}$, 
is computed using a residual connection followed by layer normalization:
\begin{equation*}
\mathbf{h}_{\mathrm{temp}}^{(i)} 
= 
\mathrm{LayerNorm}\!\big(
\mathrm{MultiHeadAttn}(\mathbf{q}, \mathbf{J}^{(i)}, \mathbf{J}^{(i)}) + \mathbf{q}
\big).
\end{equation*}

Before being fed into the multimodal fusion module, 
$\mathbf{h}_{\mathrm{temp}}^{(i)}$ is further refined by a graph learner 
to capture cross-site dependencies, resulting in the final temporal representation 
$\mathbf{f}_{\mathrm{temp}}^{(i)}$. 
Details of the graph learner are provided in Section~\ref{sec:graph_learner}.

\subsubsection{Text Encoder}

The text encoder leverages textual descriptions to summarize historical information and provide complementary semantic representations for PV power forecasting. Specifically, for each site $i$, the historical observations are converted into a descriptive text sequence $\mathbf{c}^{(i)}$ using a predefined template (see Appendix). The resulting text is then encoded using a frozen Qwen-VLM-based text encoder:
\begin{equation*}
\mathbf{f}_{\mathrm{qwen}}^{(i)}
=
\mathrm{QwenVLM}_{\mathrm{text}}\!\left(\mathbf{c}^{(i)}\right)
\in \mathbb{R}^{d_{\text{text}}}.
\end{equation*}

To further incorporate temporal information from historical numerical observations, we introduce an auxiliary temporal encoding branch aligned with the text embedding space. For site $i$, the historical multivariate time series $\mathbf{X}_{1:L}^{(i)} \in \mathbb{R}^{L \times D}$ is first projected into the feature space through a linear transformation:
\begin{equation*}
\mathbf{H}_0^{(i)} =
\mathbf{X}_{1:L}^{(i)} \mathbf{W}_{\mathrm{proj}} + \mathbf{b}_{\mathrm{proj}}
\in \mathbb{R}^{L \times d_{\text{text}}},
\end{equation*}
where $\mathbf{W}_{\mathrm{proj}} \in \mathbb{R}^{D \times d_{\text{text}}}$ and $\mathbf{b}_{\mathrm{proj}} \in \mathbb{R}^{d_{\text{text}}}$ are trainable parameters.
The projected sequence is then processed by two stacked Transformer encoder layers composed of multi-head self-attention and feed-forward networks, yielding
\begin{equation*}
\mathbf{H}^{(i)} =
\mathrm{TransformerEncoder}\!\left(\mathbf{H}_0^{(i)}\right)
=
\left(\mathbf{H}_1^{(i)}, \ldots, \mathbf{H}_L^{(i)}\right),
\quad
\mathbf{H}_t^{(i)} \in \mathbb{R}^{d_{\text{text}}}.
\end{equation*}

To summarize both recent dynamics and long-range temporal context, we adopt a weighted temporal aggregation strategy:
\begin{equation*}
\mathbf{f}_{\mathrm{aux}}^{(i)}
=
\lambda\, \mathbf{H}_L^{(i)}
+
(1-\lambda)\,
\frac{1}{L}
\sum_{t=1}^{L}
\mathbf{H}_t^{(i)},
\end{equation*}
where $\lambda \in [0,1]$ is a predefined hyperparameter controlling the relative contribution of the most recent state and the global temporal average.

Finally, the semantic representation from the Qwen-VLM text encoder and the temporally enhanced representation are fused to obtain the final text embedding:
\begin{equation*}
\mathbf{f}_{\mathrm{text}}^{(i)}
=
\gamma^{(i)} \, \mathbf{f}_{\mathrm{qwen}}^{(i)}
+
\big(1-\gamma^{(i)}\big)\, \mathbf{f}_{\mathrm{aux}}^{(i)},
\end{equation*}
where the fusion weight $\gamma^{(i)} \in [0,1]$ is computed as
\begin{equation*}
\gamma^{(i)} =
\sigma\!\left(
\mathrm{MLP}\big(
[\mathbf{f}_{\mathrm{qwen}}^{(i)};\mathbf{f}_{\mathrm{temp}}^{(i)}]
\big)
\right).
\end{equation*}

\subsubsection{Visual Encoder}

To incorporate visual context into the forecasting framework, we design a visual encoder that models the temporal dynamics of satellite imagery associated with each photovoltaic site. Each satellite image covers a localized spatial region centered at the corresponding site, capturing cloud movement and atmospheric conditions relevant to local PV generation.

Specifically, for site $i$, the satellite image at time $t$, denoted by $\mathbf{s}^{(i)}_t$, is first encoded independently using a frozen Qwen-based image encoder:
\begin{equation*}
\mathbf{u}^{(i)}_t
=
\mathrm{Qwen}_{\mathrm{image}}\!\left(\mathbf{S}^{(i)}_t\right)
\in \mathbb{R}^{d_v}.
\end{equation*}
Given the most recent $k$ satellite images, we obtain a visual feature sequence
\begin{equation*}
\mathbf{U}^{(i)} =
\left(
\mathbf{u}^{(i)}_{L-k+1}, \ldots, \mathbf{u}^{(i)}_{L}
\right)
\in \mathbb{R}^{k \times d_v}.
\end{equation*}

To capture temporal dependencies in the visual modality, the sequence $\mathbf{U}^{(i)}$ is processed by two stacked Transformer encoder layers composed of multi-head self-attention and feed-forward networks, yielding
\begin{equation*}
\mathbf{G}^{(i)} =
\mathrm{TransformerEncoder}\!\left(\mathbf{U}^{(i)}\right)
=
\left(
\mathbf{g}^{(i)}_{L-k+1}, \ldots, \mathbf{g}^{(i)}_{L}
\right),
\quad
\mathbf{g}^{(i)}_t \in \mathbb{R}^{d_v}.
\end{equation*}

To summarize both recent visual dynamics and longer-term visual context, we apply a weighted temporal aggregation strategy:
\begin{equation*}
\mathbf{f}_{\mathrm{vis}}^{(i)}
=
\lambda_v\, \mathbf{g}^{(i)}_{L}
+
(1-\lambda_v)\,
\frac{1}{k}
\sum_{t=L-k+1}^{L}
\mathbf{g}^{(i)}_{t},
\end{equation*}
where $\lambda_v \in [0,1]$ is a predefined hyperparameter controlling the relative contribution of the most recent visual state and the averaged historical visual context.

Finally, the aggregated visual representation is projected into the shared latent space via an MLP:
\begin{equation*}
\mathbf{f}_{\mathrm{visual}}^{(i)}
=
\mathrm{MLP}\!\left(
\mathbf{f}_{\mathrm{vis}}^{(i)}
\right)
\in \mathbb{R}^{d_{\mathrm{model}}},
\end{equation*}
which serves as the visual modality embedding for site $i$ in subsequent multimodal fusion and cross-site modeling.

\subsection{Multimodal Feature Fusion}
\label{sec:multimodel_feature_fusion}

For each PV site, the temporal, textual, and visual representations are integrated to form a unified multimodal embedding that jointly captures numerical dynamics, semantic context, and visual atmospheric patterns. Specifically, for site $i$, the modality-specific representations
$\mathbf{f}_{\mathrm{temp}}^{(i)}$,
$\mathbf{f}_{\mathrm{text}}^{(i)}$,
and
$\mathbf{f}_{\mathrm{visual}}^{(i)}$
are first concatenated along the feature dimension:
\begin{equation*}
\mathbf{f}_{\mathrm{concat}}^{(i)}
=
\left[
\mathbf{f}_{\mathrm{temp}}^{(i)};
\mathbf{f}_{\mathrm{text}}^{(i)};
\mathbf{f}_{\mathrm{visual}}^{(i)}
\right]
\in \mathbb{R}^{3 d_{\mathrm{model}}}.
\end{equation*}

The concatenated representation is then projected into a shared latent space through a two-layer multilayer perceptron (MLP), yielding the fused multimodal representation:
\begin{equation*}
\mathbf{f}_{\mathrm{multi}}^{(i)}
=
\mathrm{MLP}_{\mathrm{fusion}}\!\left(
\mathbf{f}_{\mathrm{concat}}^{(i)}
\right)
\in \mathbb{R}^{d_{\mathrm{model}}}.
\end{equation*}

Stacking the fused representations of all sites results in a multimodal feature matrix
\begin{equation*}
\mathbf{F}_{\mathrm{multi}}
=
\left(
\mathbf{f}_{\mathrm{multi}}^{(1)}, \ldots, \mathbf{f}_{\mathrm{multi}}^{(M)}
\right)
\in \mathbb{R}^{M \times d_{\mathrm{model}}},
\end{equation*}
which serves as the input to subsequent cross-site modeling modules.

It is worth noting that the temporal representation $\mathbf{f}_{\mathrm{temp}}^{(i)}$ is not directly obtained from the time-series encoder output. Instead, before multimodal fusion, the patch-level temporal features $\bar{\mathbf{E}}_{d,p}^{(i)}$ are further processed by a Graph Learner module to explicitly capture spatial dependencies and inter-site correlations, resulting in the final temporal embedding $\mathbf{f}_{\mathrm{temp}}^{(i)}$. The Graph Learner module will be described in detail in the next subsection.

\subsection{Cross-Site Joint Modeling}
\label{sec:cross_site_modeling}

PV power generation at geographically distributed sites exhibits strong inter-site dependencies driven by shared meteorological systems and spatially correlated cloud dynamics. Effectively capturing such dependencies is crucial for improving multi-site forecasting performance. However, different data modalities exhibit distinct characteristics in how inter-site relationships are represented after encoding. Time-series encoder operates on comparable physical measurements, so representations from different sites remain directly comparable and can be effectively enhanced via GNN message passing. In contrast, visual and textual features are produced by VLMs pretrained for semantic alignment rather than physical consistency across sites, and are therefore less suitable for cross-site graph propagation.

\begin{figure}[t!]
    \centering
    \includegraphics[width=1.0\linewidth]{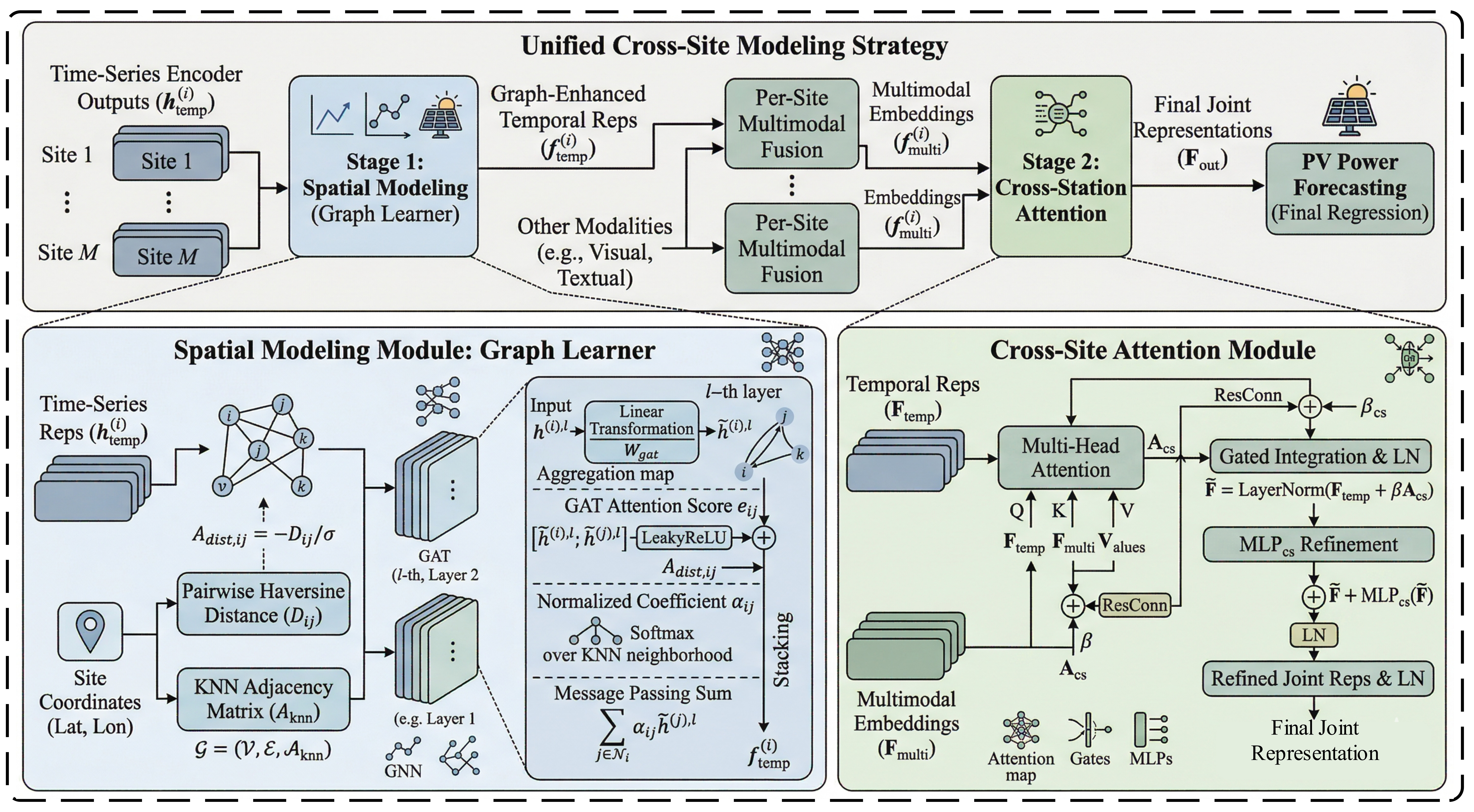}
    \caption{Overview of the proposed two-stage cross-site joint modeling strategy.}
    \label{fig:cross-site}
\end{figure}

Motivated by this observation, we adopt a two-stage cross-site joint modeling strategy. In the first stage, a graph learner module is applied exclusively to the time-series encoder outputs to capture spatial correlations among sites based on geographical proximity. In the second stage, after per-site multimodal fusion, a cross-station attention mechanism is employed to further enhance information sharing across sites by attending over fused multimodal representations. This attention-based interaction allows adaptive and data-driven cross-site fusion. In Fig.~\ref{fig:cross-site}, we illustrate the detailed architecture of the proposed two-stage cross-site joint modeling strategy.

\subsubsection{Spatial Modeling Module: Graph Learner}
\label{sec:graph_learner}

To capture spatial dependencies among photovoltaic stations, we propose a \emph{Graph Learner} that models inter-station relationships through a Graph Attention Network (GAT), enabling information propagation across sites. 
To this end, we first calculate the pairwise Haversine distances $D_{ij}$ between site $i$ and site $j$ using the latitude and longitude of each station. 
Based on the pairwise distances, we construct a $K$-nearest-neighbor (KNN) adjacency matrix $A_{\mathrm{knn}}$ by connecting each station to its $K$ nearest neighbors:
\begin{equation*}
A_{\mathrm{knn},ij} = 
\begin{cases}
1, & \text{if $j$ is among the $K$ nearest neighbors of $i$},\\
0, & \text{otherwise}.
\end{cases}
\end{equation*}

The resulting KNN graph is denoted as $\mathcal{G} = (\mathcal{V}, \mathcal{E}, A_{\mathrm{knn}})$, 
where $\mathcal{V}$ represents the set of PV sites and 
$\mathcal{E}$ denotes the edge set defined by $A_{\mathrm{knn}}$. 
To incorporate distance information into the GAT, we further define a distance-based weight matrix
\begin{equation*}
A_{\mathrm{dist},ij} = -\frac{D_{ij}}{\sigma},
\end{equation*}
where $\sigma$ is set to the median of all pairwise distances.

Based on $A_{\mathrm{dist}}$ and $A_{\mathrm{knn}}$, we define the graph attention layer of the proposed GAT.
At the $l$-th graph attention layer, the feature representation of station $i$ is denoted by $\mathbf{h}^{(i),l}$, where $\mathbf{h}^{(i),0}=\mathbf{h}_{\mathrm{temp}}^{(i)}$.
First, a linear transformation is applied:
\begin{equation*}
\tilde{\mathbf{h}}^{(i),l}
=
\mathbf{W}_{\mathrm{gat}}
\mathbf{h}^{(i),l}.
\end{equation*}

Next, edge-wise attention scores are computed, whose values are dynamically adjusted based on the features of stations $i$ and $j$ while incorporating the distance-based weights $A_{\mathrm{dist},ij}$:
\begin{equation*}
e_{ij}
=
\mathrm{LeakyReLU}
\!\left(
\mathbf{a}^{\top}
\left[
\tilde{\mathbf{h}}^{(i),l};
\tilde{\mathbf{h}}^{(j),l}
\right]
\right)
+
A_{\mathrm{dist},ij},
\end{equation*}
where $\mathbf{a}$ is a learnable attention vector.
The attention coefficients are then normalized over the neighborhood defined by the KNN graph:
\begin{equation*}
\alpha_{ij}
=
\frac{\exp(e_{ij})}
{\sum_{k \in \mathcal{N}_i} \exp(e_{ik})},
\end{equation*}
where $\mathcal{N}_i$ denotes the set of neighbors of station $i$ in the KNN graph.
Finally, message passing is performed by aggregating the transformed features of neighboring stations:
\begin{equation*}
\mathbf{h}^{(i),l+1}
=
\sum_{j \in \mathcal{N}_i}
\alpha_{ij}\,
\tilde{\mathbf{h}}^{(j),l}.
\end{equation*}

After stacking two graph attention layers, the resulting representation is taken as the final temporal embedding for station $i$, i.e.,  $\mathbf{f}_{\mathrm{temp}}^{(i)}$.

\subsubsection{Cross-Site Attention}

After graph-based spatial modeling of numerical time-series features and per-site multimodal fusion, we further enhance inter-site information exchange through a cross-site attention mechanism. This module allows each site to adaptively aggregate relevant information from other sites in a data-driven manner, without imposing explicit structural constraints on heterogeneous multimodal embeddings.

Recall that the graph learner produces a graph-enhanced temporal representation
$\mathbf{f}_{\mathrm{temp}}^{(i)} \in \mathbb{R}^{d_{\mathrm{model}}}$
for each site $i$, while multimodal fusion yields the per-site multimodal embedding
$\mathbf{f}_{\mathrm{multi}}^{(i)} \in \mathbb{R}^{d_{\mathrm{model}}}$.
Stacking the features of all $M$ sites results in
\begin{equation*}
\mathbf{F}_{\mathrm{temp}} \in \mathbb{R}^{M \times d_{\mathrm{model}}},
\qquad
\mathbf{F}_{\mathrm{multi}} \in \mathbb{R}^{M \times d_{\mathrm{model}}},
\end{equation*}
which are used as inputs for cross-site interaction modeling.

To propagate multimodal information across sites while preserving the structural role of temporal features, we apply multi-head attention by treating temporal features as queries and multimodal features as keys and values:
\begin{equation*}
\mathbf{A}_{\mathrm{cs}}
=
\mathrm{MultiHeadAttn}\!\left(
\mathbf{F}_{\mathrm{temp}},
\mathbf{F}_{\mathrm{multi}},
\mathbf{F}_{\mathrm{multi}}
\right),
\end{equation*}
where each site selectively attends to multimodal representations from other sites according to their relevance to its temporal dynamics.

The attention output is integrated with the original temporal features through a residual connection with a learnable scaling factor:
\begin{equation*}
\tilde{\mathbf{F}}
=
\mathrm{LayerNorm}\!\left(
\mathbf{F}_{\mathrm{temp}} + \beta\, \mathbf{A}_{\mathrm{cs}}
\right),
\end{equation*}
where $\beta$ is a trainable scalar that controls the strength of cross-site information aggregation.

To further refine the representations, a position-wise multilayer perceptron specific to cross-site modeling is applied, followed by another residual connection and layer normalization:
\begin{equation*}
\mathbf{F}_{\mathrm{out}}
=
\mathrm{LayerNorm}\!\left(
\tilde{\mathbf{F}} + \mathrm{MLP}_{\mathrm{cs}}(\tilde{\mathbf{F}})
\right).
\end{equation*}

The resulting representation
$\mathbf{F}_{\mathrm{out}} \in \mathbb{R}^{M \times d_{\mathrm{model}}}$
encodes cross-site dependencies by enriching each site’s temporal features with selectively attended multimodal information from other sites, and is used as the final joint representation for downstream photovoltaic power forecasting.

\subsection{Final Prediction}

The final representation obtained from cross-site attention,
$\mathbf{f}_{\mathrm{out}}^{(i)}$,
is fed into a MLP to generate the $T$-step-ahead PV power forecast for site $i$:
\begin{equation*}
\mathbf{p}_{\mathrm{out}}^{(i)} = \mathrm{MLP}\left(\mathbf{f}_{\mathrm{out}}^{(i)}\right),
\quad
\mathbf{p}_{\mathrm{out}}^{(i)} \in \mathbb{R}^{T}.
\end{equation*}

Since temporal information constitutes the most fundamental signal for PV power forecasting, we additionally produce an auxiliary prediction based solely on the temporal representation
$\mathbf{h}_{\mathrm{temp}}^{(i)}$:
\begin{equation*}
\mathbf{p}_{\mathrm{temp}}^{(i)} = \mathrm{MLP}\left(\mathbf{h}_{\mathrm{temp}}^{(i)}\right),
\quad
\mathbf{p}_{\mathrm{temp}}^{(i)} \in \mathbb{R}^{T}.
\end{equation*}

The two predictions are adaptively fused via a gating mechanism. The fusion weight is computed as
\begin{equation*}
\theta^{(i)} =
\sigma\left(
\mathrm{MLP}\left(
[\mathbf{f}_{\mathrm{out}}^{(i)};\mathbf{h}_{\mathrm{temp}}^{(i)}]
\right)
\right),
\end{equation*}
and the final prediction is given by
\begin{equation*}
\mathbf{p}^{(i)} =
\theta^{(i)}\, \mathbf{p}_{\mathrm{out}}^{(i)}
+
\left(1-\theta^{(i)}\right)\, \mathbf{p}_{\mathrm{temp}}^{(i)}.
\end{equation*}

\section{Experimental Results}

\subsection{Experiment Setting}

To evaluate the effectiveness of Solar-VLM, we conduct experiments on a real-world PV dataset collected from eight PV power stations located in Hebei Province, China. The dataset includes both historical numerical measurements and satellite cloud imagery. The historical data consist of three components: numerical weather prediction (NWP) variables, local meteorological data (LMD), and PV power output. Specifically, the NWP variables include global horizontal irradiance, direct irradiance, temperature, humidity, wind speed, wind direction, and atmospheric pressure, while the LMD variables contain the same meteorological measurements except for direct irradiance. Together with the PV power output, a total of 14 features are used in the experiments. All numerical observations are recorded at 15-minute intervals. The satellite cloud images are extracted based on a region of interest centered on each station, with a spatial resolution of $128 \times 128$ pixels, covering the surrounding area of each PV site.

To comprehensively evaluate the proposed model, we compare Solar-VLM with several representative baselines. These include modern time-series forecasting models trained from scratch on the target PV dataset, such as Informer~\cite{Informer}, FEDformer~\cite{FEDformer}, TimesNet~\cite{timesnet}, and TimeLLM~\cite{TimeLLM}. In addition, we include SUNSET~\cite{SUNSET}, a CNN-based model specifically designed for PV power forecasting, as well as Time-VLM~\cite{Time_VLM}, a vision-language model developed for time-series forecasting tasks.

For forecasting, the model uses the most recent three days of historical data (288 time steps) as input to predict PV power output over different prediction horizons $T = 3, 6, 12, 24, 48, 96$. The complete hyperparameter configuration is reported in Table~\ref{tab:hyperparameters}. To evaluate forecasting performance, mean squared error (MSE), mean absolute error (MAE), and the coefficient of determination ($R^2$) are adopted as evaluation metrics. MSE measures the average squared difference between predicted and true values and places greater emphasis on large errors, while MAE computes the average absolute difference between predictions and actual values. In addition, $R^2$ reflects the proportion of variance in the observed data that can be explained by the model, indicating the overall goodness of fit.

\begin{table}[t!]
    \centering
    \caption{Model Hyperparameters}
    \label{tab:hyperparameters}
    \begin{tabular}{>{\raggedright\arraybackslash}p{4cm} >{\raggedright\arraybackslash}p{6cm} c >{\raggedright\arraybackslash}p{3cm}}
        \toprule
        \textbf{Module} & \textbf{Hyperparameter} & \textbf{Symbol} & \textbf{Value} \\
        \midrule
        \multirow{5}{*}{\textbf{Time Series Encoder}} 
        & Patch length & $l_p$ & 10 \\
        & Patch stride & $s_p$ & 8 \\
        & Embedding dimension & $d_{\mathrm{model}}$ & 128 \\
        & Memory bank size & $N_m$ & 100 \\
        & Number of retrieved memories & $k$ & 5 \\
        \midrule
        \multirow{2}{*}{\textbf{Text Encoder}} 
        & Qwen-VLM output dimension & $d_{\text{text}}$ & 2048 \\
        & Temporal aggregation weight & $\lambda$ & 0.6 \\
        \midrule
        \multirow{3}{*}{\textbf{Visual Encoder}} 
        & Qwen image encoder dimension & $d_v$ & 2048 \\
        & Number of satellite images & $k$ & 8 \\
        & Visual temporal aggregation weight & $\lambda_v$ & 0.7 \\
        \midrule
        \multirow{1}{*}{\textbf{Graph Learner}} 
        & KNN neighbors & $K$ & 5 \\
        \midrule
        \multirow{3}{*}{\textbf{Training Settings}} 
        & Batch size & - & 16 \\
        & Learning rate & - & $1\times 10^{-4}$ \\
        & Number of epochs & - & 50 \\
        \bottomrule
    \end{tabular}
\end{table}

\subsection{PV Forecasting Results}

\begin{table}[h!]
\centering
\caption{Performance Comparison of Different Models}
\label{tab:results}
\resizebox{\textwidth}{!}{
\begin{tabular}{c|ccc|ccc|ccc|ccc|ccc|ccc}
\toprule
\multirow{2}{*}{\textbf{Model}} 
& \multicolumn{3}{c|}{$T=6$}
& \multicolumn{3}{c|}{$T=12$}
& \multicolumn{3}{c|}{$T=24$}
& \multicolumn{3}{c|}{$T=36$}
& \multicolumn{3}{c|}{$T=48$}
& \multicolumn{3}{c}{$T=96$} \\
\cmidrule(lr){2-4}
\cmidrule(lr){5-7}
\cmidrule(lr){8-10}
\cmidrule(lr){11-13}
\cmidrule(lr){14-16}
\cmidrule(lr){17-19}
& MSE & MAE & $R^2$
& MSE & MAE & $R^2$ 
& MSE & MAE & $R^2$ 
& MSE & MAE & $R^2$
& MSE & MAE & $R^2$ 
& MSE & MAE & $R^2$ \\
\midrule
LSTM     
& 0.175 & \underline{0.231} & 0.932
& 0.279 & 0.299 & 0.891
& 0.703 & 0.508 & 0.727
& 0.714 & 0.501 & 0.723
& 0.564 & 0.461 & 0.781
& 0.595 & 0.449 & 0.770 \\

Informer  
& 0.937 & 0.609 & 0.635
& 0.594 & 0.494 & 0.769
& 0.427 & 0.361 & 0.834
& 0.494 & 0.425 & 0.808
& 0.452 & 0.381 & 0.825
& 0.495 & 0.392 & 0.808 \\

FEDformer 
& 0.383 & 0.413 & 0.851
& 0.427 & 0.474 & 0.834
& 0.458 & 0.481 & 0.822
& 0.450 & 0.484 & 0.825
& 0.485 & 0.495 & 0.812
& 0.534 & 0.567 & 0.793 \\

TimesNet  
& 0.228 & 0.298 & 0.911
& 0.295 & 0.377 & 0.885
& 0.407 & 0.412 & 0.842
& 0.428 & 0.425 & 0.834
& 0.443 & 0.452 & 0.828
& 0.681 & 0.534 & 0.736 \\

TimeLLM   
& 0.287 & 0.329 & 0.888
& 0.335 & 0.349 & 0.869
& 0.378 & 0.388 & 0.853
& 0.435 & 0.422 & 0.831
& 0.418 & 0.396 & 0.838
& 0.447 & 0.420 & 0.827 \\

SUNSET    
& 0.276 & 0.295 & 0.893
& 0.352 & 0.350 & 0.864
& 0.342 & 0.357 & 0.868
& 0.380 & 0.375 & 0.855
& 0.480 & 0.439 & 0.816
& 0.448 & 0.447 & 0.827 \\

TimeVLM   
& \underline{0.164} & 0.243 & \underline{0.936}
& \underline{0.223} & \underline{0.273} & \underline{0.913}
& \underline{0.304} & \underline{0.317} & \underline{0.882}
& \underline{0.339} & \underline{0.331} & \underline{0.868}
& \underline{0.361} & \underline{0.337} & \underline{0.860}
& \underline{0.382} & \underline{0.357} & \underline{0.852} \\

\textbf{Solar-VLM} 
& \textbf{0.160} & \textbf{0.217} & \textbf{0.940}
& \textbf{0.166} & \textbf{0.216} & \textbf{0.938}
& \textbf{0.186} & \textbf{0.221} & \textbf{0.930}
& \textbf{0.202} & \textbf{0.256} & \textbf{0.924}
& \textbf{0.232} & \textbf{0.286} & \textbf{0.913}
& \textbf{0.309} & \textbf{0.349} & \textbf{0.883} \\

\bottomrule
\end{tabular}
}
\end{table}

As shown in table~\ref{tab:results}, Solar-VLM achieves the best performance in the vast majority of settings. The improvements are particularly significant when the forecasting horizons are $T=24$ and $T=48$. Specifically, when $T=24$, Solar-VLM reduces the MSE by 3.5\% and the MAE by 10.5\% compared with the second-best model. When $T=48$, the improvements become even more pronounced, with MSE reduced by 8.5\% and MAE reduced by 18.3\%.
These results demonstrate the effectiveness of integrating temporal, visual, and textual modalities, as well as performing joint spatio-temporal modeling across multiple PV stations. By leveraging complementary information from multiple modalities and capturing inter-station dependencies, Solar-VLM achieves superior forecasting performance.

\begin{figure}[t!]
    \centering
    \includegraphics[width=1.0\linewidth]{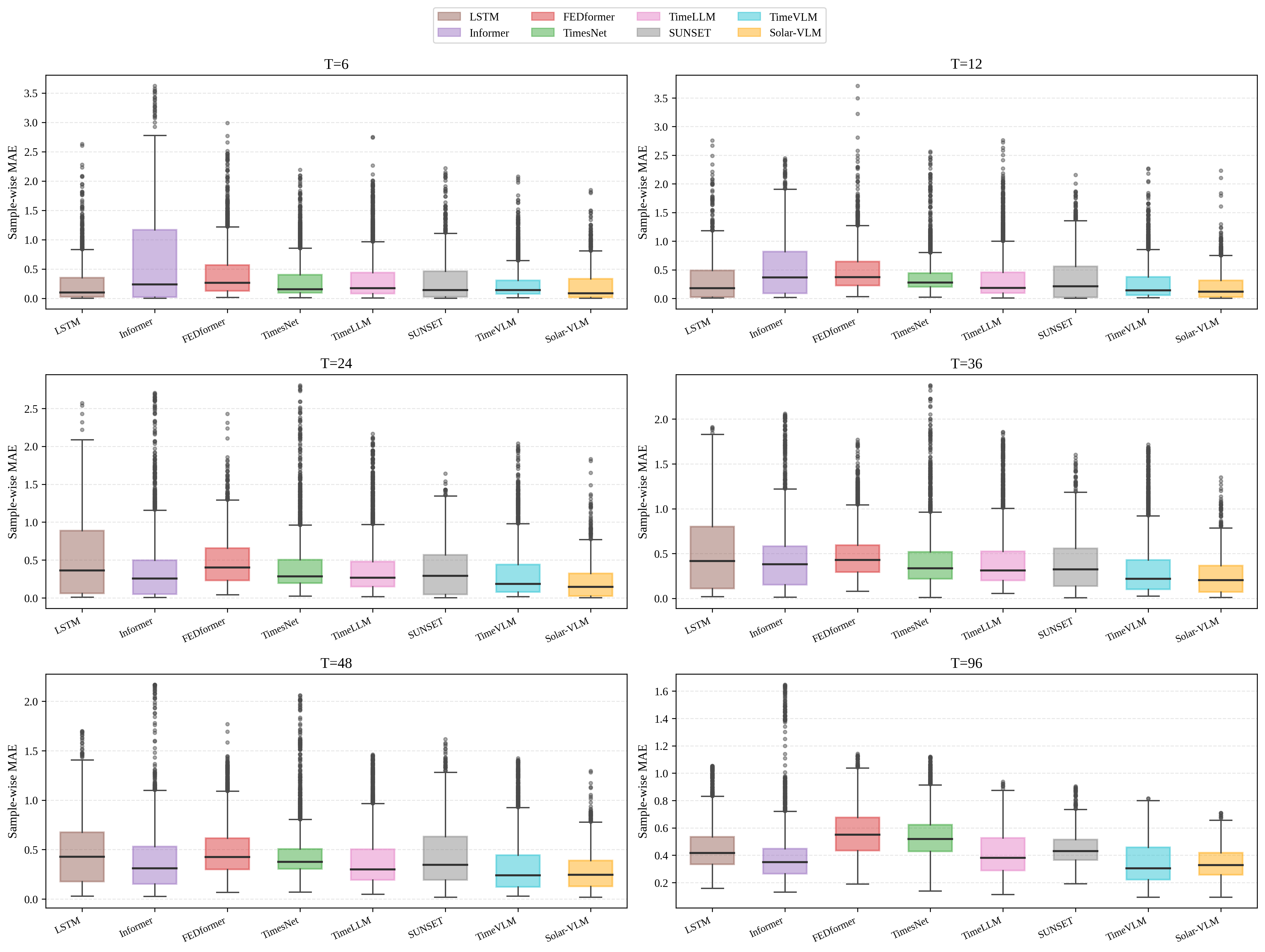}
    \caption{MAE boxplots for different models at different forecasting horizons.
    }
    \label{fig:boxplot}
\end{figure}

Figure~\ref{fig:boxplot} further presents the MAE boxplots of all compared models across different forecasting horizons. In most cases, the proposed Solar-VLM achieves not only the lowest mean MAE but also one of the smallest variances, which further demonstrates the excellent stability of its prediction performance.

\begin{figure}[t!]
    \centering
    \includegraphics[width=1.0\linewidth]{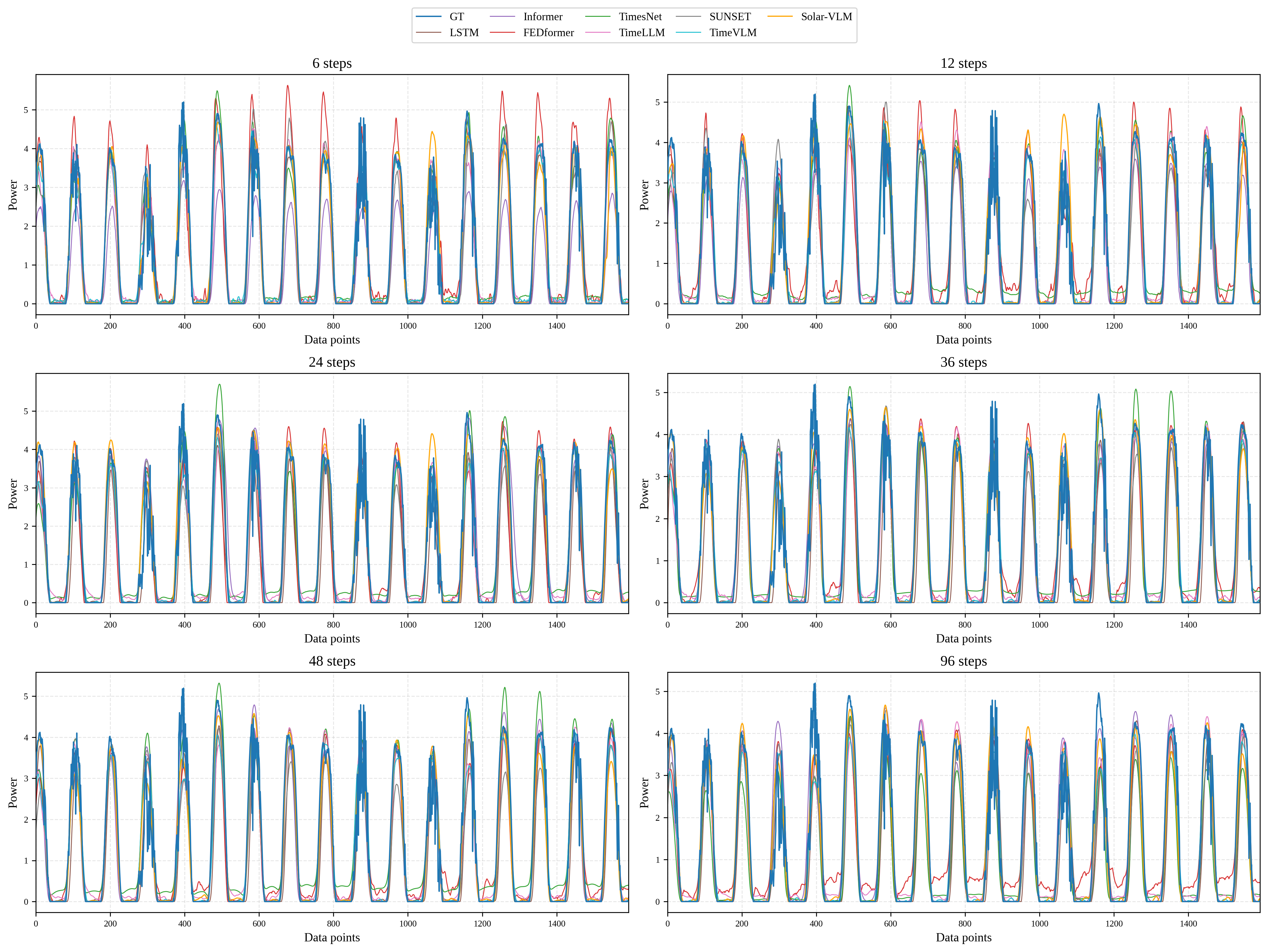}
    \caption{PV power prediction curves of compared models against the ground truth on the test set at different forecasting horizons.
    }
    \label{fig:prediction_curve}
\end{figure}

To intuitively compare the prediction performance of different models on the test set, Figure \ref{fig:prediction_curve} presents the PV power forecasting curves of all compared models against the ground truth across various forecasting horizons. As shown in the figure, the proposed Solar-VLM achieves the smallest overall deviation from the ground truth curves.

\subsection{Ablation Study}

To evaluate the contribution of each component in Solar-VLM, we conduct an ablation study by systematically removing key modules from the full model. Specifically, we remove the text encoder, visual encoder, Graph Learner, and cross-site attention module, respectively. All variants are trained and evaluated under the forecasting horizon $T=48$, and the corresponding MSE and MAE scores are reported. 
Since temporal information constitutes the fundamental signal for PV power forecasting, we additionally evaluate a variant that retains only the time-series encoder. The experimental results are presented in Table~\ref{tab:ablation_study}.

\begin{table}[h!]
\centering
\caption{Ablation Study on the Proposed Model}
\label{tab:ablation_study}
\begin{tabular}{c|ccc}
\toprule
\textbf{Setting} & \textbf{MSE} & \textbf{MAE} & $R^2$ \\
\midrule
\textbf{Full}                        & \textbf{0.207} & \textbf{0.254} & \textbf{0.888} \\
Time-Series Encoder Only    & 0.256 & 0.296 & 0.862  \\
Without Text Encoder        & 0.233 & 0.274 & 0.874 \\
Without Visual Encoder      & 0.211 & 0.256 & 0.886 \\
Without Graph Leaner        & 0.218 & 0.261 & 0.882 \\
Without Cross-Site Attention & 0.301 & 0.296 & 0.862 \\
\bottomrule
\end{tabular}
\end{table}

As shown in Table~\ref{tab:ablation_study}, removing any individual module leads to performance degradation compared with the full model, indicating that each component contributes to the forecasting task. When only the time-series encoder is retained, the performance drops noticeably, suggesting that although temporal dynamics provide the fundamental predictive signal, incorporating additional modalities and spatial modeling further improves forecasting accuracy. Among all variants, removing the cross-site attention module results in the most significant deterioration, highlighting the importance of adaptive inter-site information aggregation in multi-site PV forecasting. In addition, removing the text or visual encoder also leads to consistent performance drops, confirming that semantic descriptions and satellite imagery provide complementary information beyond numerical measurements.

\begin{figure}[t!]
    \centering
    \includegraphics[width=1.0\linewidth]{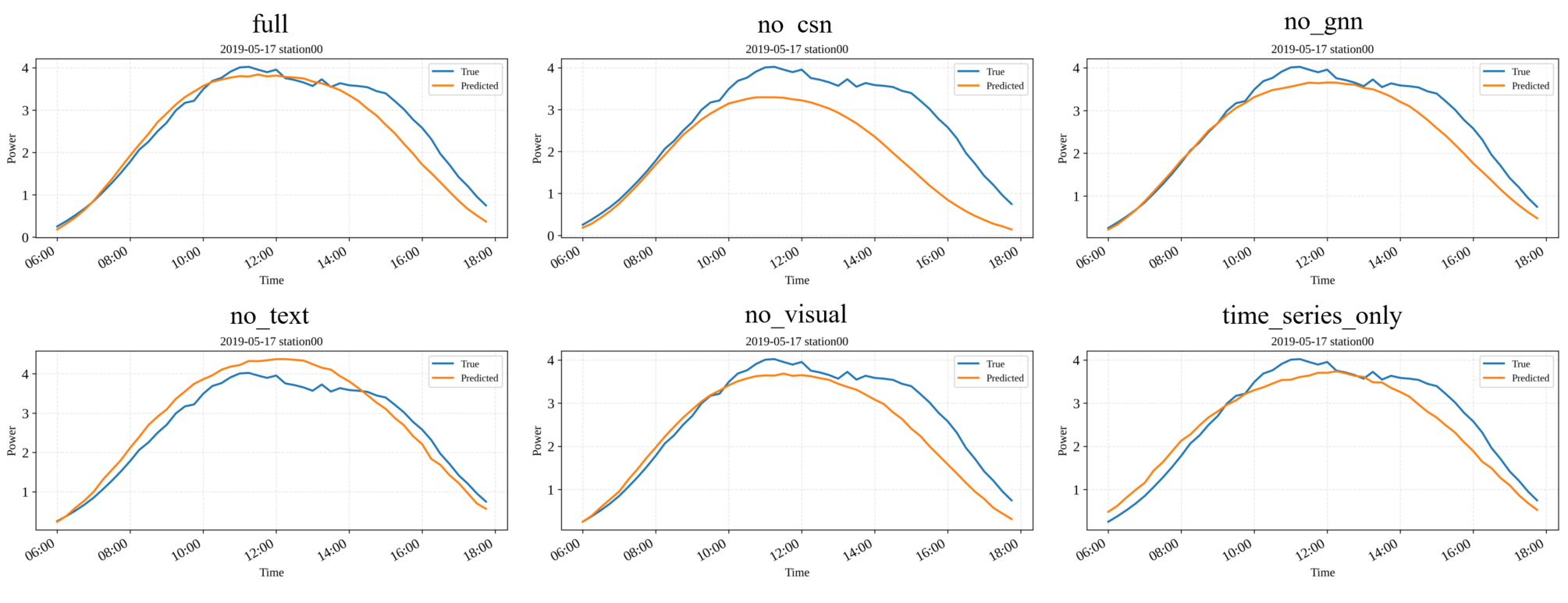}
    \caption{Forecasting results of different ablation variants on a representative cloudy day.
    }
    \label{fig:ablation1}
\end{figure}

\begin{figure}[t!]

    \centering
    \includegraphics[width=1.0\linewidth]{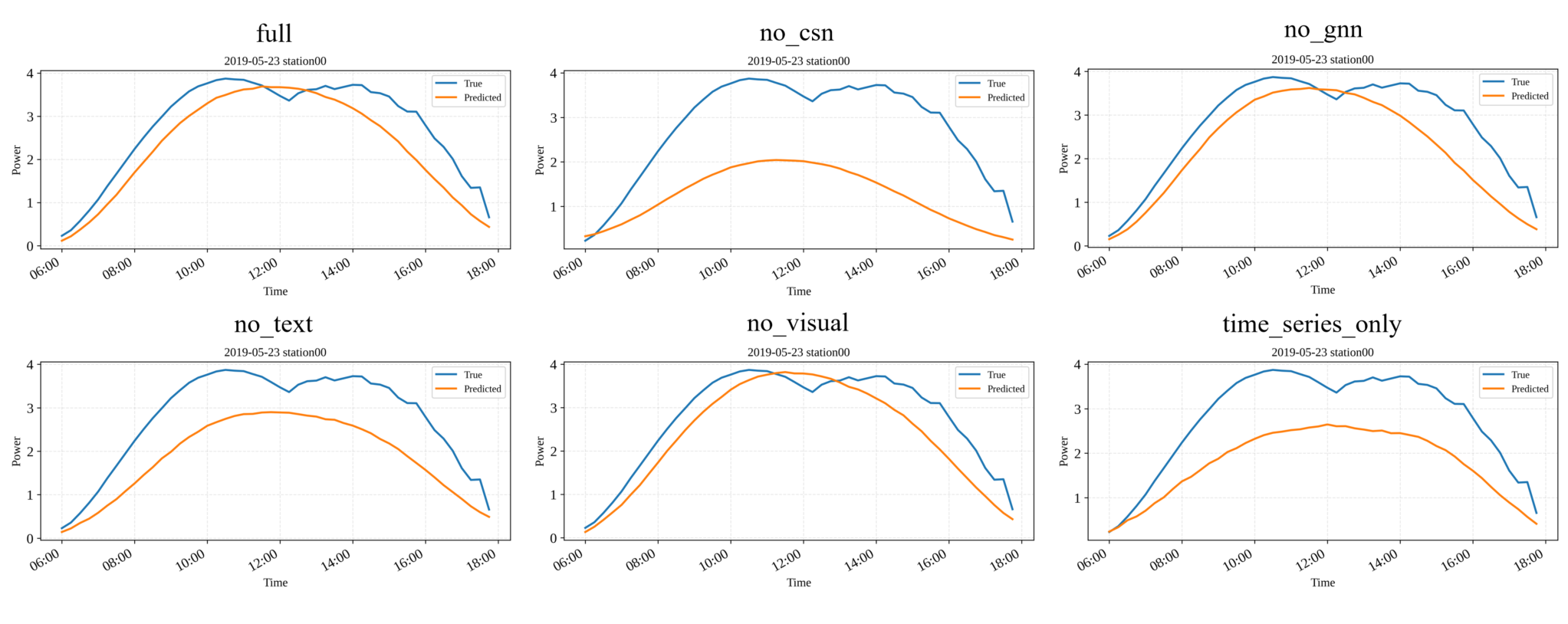}
    \caption{Forecasting results of different ablation variants on a representative sunny day.
    }
    \label{fig:ablation2}
\end{figure}

Fig.~\ref{fig:ablation1} and Fig.~\ref{fig:ablation2} further present qualitative comparisons on a representative cloudy day and a sunny day, respectively. As observed, the predictions of the full model follow the ground-truth PV curve more closely, while the ablated variants exhibit larger deviations, further demonstrating the effectiveness of the proposed framework.

\subsection{Sensitivity Analysis of Key Parameters}

\begin{figure}[t!]
    \centering
    \includegraphics[width=1.0\linewidth]{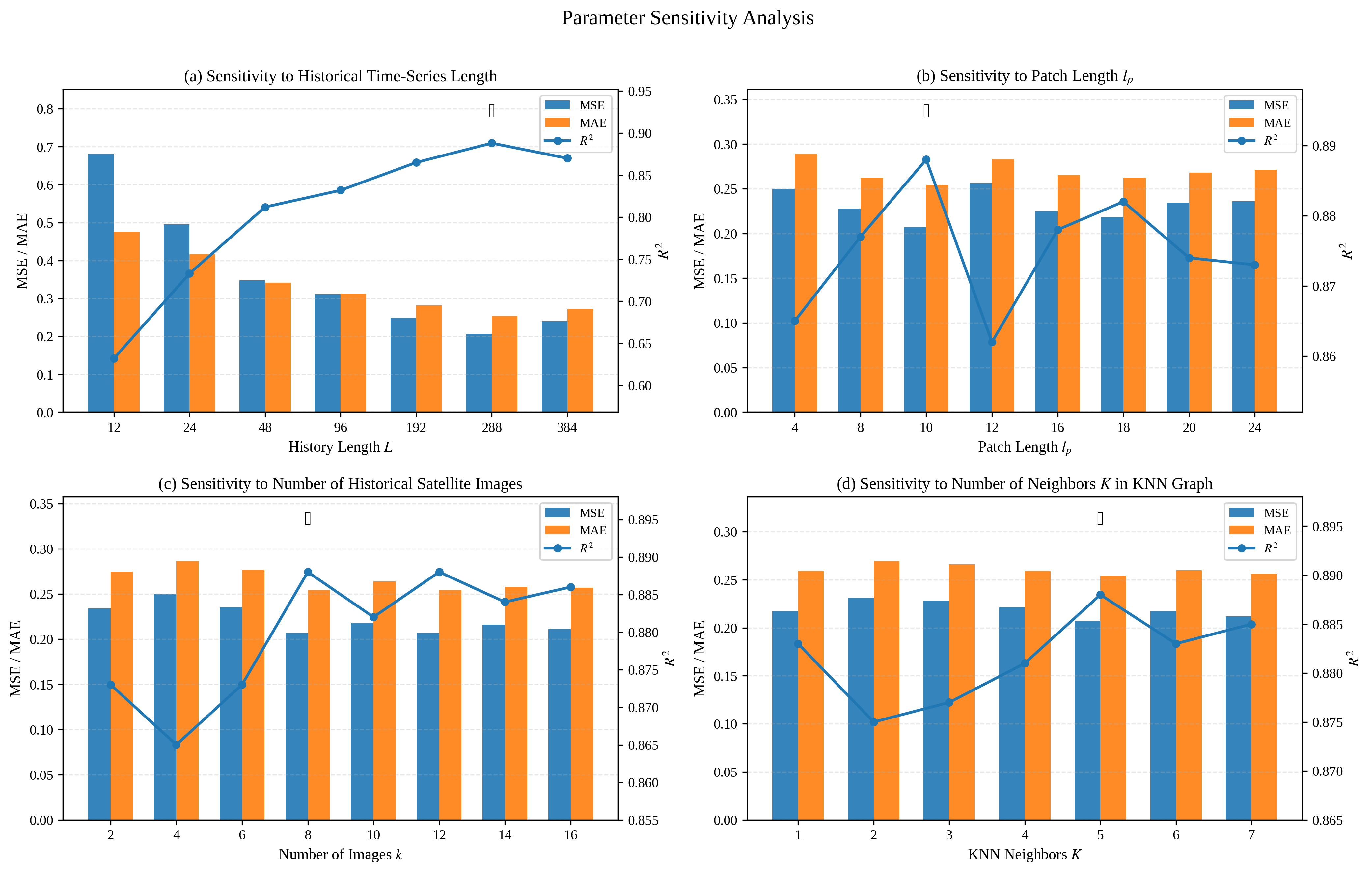}
    \caption{Sensitivity analysis of key hyperparameters. 
    }
    \label{fig:parameter_sensitivity}
\end{figure}

To evaluate the robustness of the proposed framework and understand the influence of key design choices, we conduct sensitivity analysis on several critical hyperparameters. Specifically, we examine:
\begin{itemize}
\item The length of historical numerical observations;
\item The patch length in the time-series encoder;
\item The number of historical satellite images;
\item The number of neighbors $K$ in the KNN graph of the Graph Learner. 
\end{itemize}
Unless otherwise specified, all other hyperparameters remain fixed as reported in Table~\ref{tab:hyperparameters}. The forecasting horizon is set to $T=48$ for all experiments. To provide an intuitive understanding of how each hyperparameter affects model performance, we visualize the sensitivity trends in Fig.~\ref{fig:parameter_sensitivity}, while the exact numerical results are reported in Tables~\ref{tab:sensitivity_history_length}–\ref{tab:sensitivity_knn_k}.

\subsubsection{Effect of Historical Time-Series Length}

The length of historical numerical observations determines the amount of temporal context available for PV power forecasting. To investigate its impact on prediction performance, we vary the input sequence length $L$ while keeping other settings fixed. The results are summarized in Table~\ref{tab:sensitivity_history_length}.

\begin{table}[h!]
\centering
\caption{Sensitivity to Historical Time-Series Length}
\label{tab:sensitivity_history_length}
\begin{tabular}{c|ccc}
\toprule
\textbf{History Length $L$} & \textbf{MSE} & \textbf{MAE} & $R^2$ \\
\midrule
12  & 0.681 & 0.476 & 0.632 \\
24  & 0.495 & 0.416 & 0.733 \\
48  & 0.348 & 0.342 & 0.812 \\
96  & 0.311 & 0.312 & 0.832 \\
192  & 0.249 & 0.282 & 0.865 \\
\textbf{288}  & \textbf{0.207} & \textbf{0.254} & \textbf{0.888} \\
384  & 0.240 & 0.272 & 0.870 \\
\bottomrule
\end{tabular}
\end{table}

As $L$ increases, the forecasting accuracy first improves and then deteriorates. When $L$ is too small, the model has access to limited historical information, which restricts its ability to capture sufficient temporal patterns for reliable forecasting. Conversely, when $L$ becomes excessively large, earlier observations may contribute limited useful information while increasing model complexity and training difficulty. The inclusion of overly long historical sequences may also introduce noise and redundant patterns that interfere with effective learning.
Considering both prediction accuracy and model stability, we select $L=192$ as the optimal historical input length.

\subsubsection{Effect of Patch Length in Time-Series Encoder}

The patch length $l_p$ determines the granularity of temporal segmentation in the patch-based time-series encoder. To investigate its impact on forecasting performance, we vary $l_p$ while keeping other hyperparameters unchanged. The corresponding results are reported in Table~\ref{tab:sensitivity_patch_length}.

\begin{table}[h!]
\centering
\caption{Sensitivity to Patch Length $l_p$}
\label{tab:sensitivity_patch_length}
\begin{tabular}{c|ccc}
\toprule
\textbf{Patch Length $l_p$} & \textbf{MSE} & \textbf{MAE} & $R^2$ \\
\midrule
4   & 0.250 & 0.289 & 0.865 \\
8   & 0.228 & 0.262 & 0.877 \\
\textbf{10}   & \textbf{0.207} & \textbf{0.254} & \textbf{0.888} \\
12   & 0.256 & 0.283 & 0.862 \\
16  & 0.225 & 0.265 & 0.878 \\
18  & 0.218 & 0.262 & 0.882 \\
20  & 0.234 & 0.268 & 0.874 \\
24  & 0.236 & 0.271 & 0.873 \\
\bottomrule
\end{tabular}
\end{table}

As $l_p$ increases, the forecasting accuracy first improves and then declines. When $l_p$ is too small, each patch contains only a limited number of observations, which restricts the encoder’s ability to effectively capture meaningful local temporal patterns. In this case, the advantages of patch-based modeling cannot be fully exploited. On the other hand, when $l_p$ becomes excessively large, each patch spans an overly long time interval, which weakens the model’s capacity to focus on localized temporal characteristics and may blur fine-grained dynamics within the sequence.
Balancing local pattern modeling capability and temporal representation granularity, we select $l_p=8$ as the optimal patch length in our framework.

\subsubsection{Effect of Historical Satellite Image Sequence Length}

To evaluate the impact of visual temporal context, we vary the number of recent satellite images $k$ used in the visual encoder while keeping other settings unchanged. The results are summarized in Table~\ref{tab:sensitivity_image_length}. As $k$ increases, the forecasting accuracy first improves and then declines. This trend is consistent with the observations for the historical time-series length $L$. When $k$ is too small, the available visual information is insufficient, limiting the model’s ability to capture cloud evolution patterns and other dynamic meteorological cues from consecutive satellite images. In contrast, when $k$ becomes excessively large, earlier frames may provide limited additional predictive value while increasing training complexity and introducing potential redundancy. Therefore, an appropriate visual sequence length is necessary to balance information richness and modeling robustness.

\begin{table}[h!]
\centering
\caption{Sensitivity to Number of Historical Satellite Images}
\label{tab:sensitivity_image_length}
\begin{tabular}{c|ccc}
\toprule
\textbf{Number of Images $k$} & \textbf{MSE} & \textbf{MAE} & $R^2$ \\
\midrule
2  & 0.234 & 0.275 & 0.873 \\
4  & 0.250 & 0.286 & 0.865 \\
6  & 0.235 & 0.277 & 0.873 \\
\textbf{8}  & \textbf{0.207} & \textbf{0.254} & \textbf{0.888} \\
10  & 0.218 & 0.264 & 0.882 \\
12  & 0.207 & 0.254 & 0.888 \\
14  & 0.216 & 0.258 & 0.884 \\
16 & 0.211 & 0.257 & 0.886 \\
\bottomrule
\end{tabular}
\end{table}

\subsubsection{Effect of $K$ in the KNN Graph}

The parameter $K$ determines the number of neighboring stations considered in the KNN graph constructed by the Graph Learner. Table~\ref{tab:sensitivity_knn_k} presents the model performance under different values of $K$, which exhibits a similar trend of first improving and then deteriorating. When $K$ is too small, information exchange is restricted to a very limited local neighborhood, preventing the model from fully capturing spatial dependencies among geographically correlated stations. Conversely, when $K$ becomes excessively large, the constructed graph gradually loses its ability to characterize meaningful proximity relationships, as connections are introduced between weakly related stations. Empirically, $K=4$ achieves the best performance and is therefore adopted in our final model.

\begin{table}[h!]
\centering
\caption{Sensitivity to Number of Neighbors $K$ in KNN Graph}
\label{tab:sensitivity_knn_k}
\begin{tabular}{c|ccc}
\toprule
\textbf{KNN Neighbors $K$} & \textbf{MSE} & \textbf{MAE} & $R^2$ \\
\midrule
1  & 0.217 & 0.259 & 0.883 \\
2  & 0.231 & 0.269 & 0.875 \\
3  & 0.228 & 0.266 & 0.877 \\
4  & 0.221 & 0.259 & 0.881 \\
\textbf{5}  & \textbf{0.207} & \textbf{0.254} & \textbf{0.888} \\
6  & 0.217 & 0.260 & 0.883 \\
7  & 0.212 & 0.256 & 0.885 \\
\bottomrule
\end{tabular}
\end{table}

\section{Conclusion}

We present Solar-VLM, a multimodal multi-site framework for PV power forecasting. By integrating temporal, visual, and textual modalities and leveraging the pretrained Qwen VLM, the proposed model benefits from the large-scale representational capacity of foundation models. In addition, a graph neural network together with a cross-site attention mechanism is employed to capture spatial dependencies among geographically distributed PV stations.
Extensive experiments demonstrate that Solar-VLM consistently outperforms existing methods across multiple forecasting horizons and evaluation metrics. Furthermore, the ablation study verifies the effectiveness of each component in the proposed architecture.
In future work, we plan to further explore the capability of large vision-language models for PV forecasting, particularly in few-shot and low-data scenarios.

\section*{Acknowledgment}

This work was supported by Natural Science Foundation of Hebei Province (G2025502003, Research on Multi-time and Spatial Scale Power Prediction of Distributed Photovoltaics Based on Generative Artificial Intelligence).

\section*{Data availability}

Data will be made available on request.

\appendix
\section{Prompt Design of Text Encoder}

We adopt a structured template to transform multivariate time series context into a compact natural language representation. For each PV station, the prompt is constructed as a sequence of key-value pairs concatenated using the delimiter ``|''. As shown in Table~\ref{tab:prompt_field}, the fields are organized into four semantic groups, covering task configuration, spatiotemporal context, power generation status, and meteorological conditions, enabling a comprehensive yet concise description of the forecasting scenario.

\begin{table*}[t!]
\centering
\caption{Prompt Field Definitions.}
\label{tab:prompt_field}
\resizebox{\textwidth}{!}{
\begin{tabular}{lll}
\toprule
\textbf{Category} & \textbf{Field} & \textbf{Description} \\
\midrule

\multirow{7}{*}{Task Configuration}
& task & Fixed task identifier \\
& target & Target variable for forecasting \\
& output & Output format specification \\
& forecast\_steps & Number of steps to forecast \\
& step\_minutes & Temporal resolution in minutes \\
& context\_short\_steps & Short-term context window length \\
& context\_long\_steps & Long-term context window length \\

\midrule
\multirow{5}{*}{Spatiotemporal Context}
& site\_latitude\_band & Latitude range of the station \\
& site\_longitude\_band & Longitude range of the station \\
& season & Season of the year \\
& time\_of\_day & Time period within the day \\
& solar\_elevation & Solar elevation angle level \\

\midrule
\multirow{4}{*}{Power Generation Status}
& power\_last\_bin & Latest power output level \\
& power\_trend\_short & Power trend over short-term context \\
& power\_trend\_long & Power trend over long-term context \\
& power\_variability\_short & Power variability over short-term context \\

\midrule
\multirow{12}{*}{Meteorological Conditions}
& irradiance\_last\_bin & Latest irradiance level \\
& irradiance\_trend\_short & Irradiance trend over short-term context \\
& irradiance\_trend\_long & Irradiance trend over long-term context \\
& irradiance\_variability\_short & Irradiance variability over short-term context \\
& temperature\_last\_bin & Latest temperature level \\
& humidity\_last\_bin & Latest humidity level \\
& wind\_speed\_last\_bin & Latest wind speed level \\
& pressure\_last\_bin & Latest pressure level \\
& cloud\_indicator & Cloud condition \\
& irradiance\_power\_coherence\_short & Short-term correlation between irradiance and power \\
& irradiance\_to\_power\_response\_delay & Delay of power response to irradiance changes \\
& forecast\_observation\_irradiance\_gap & Irradiance difference between forecast and observation \\

\bottomrule
\end{tabular}
}
\end{table*}

An example prompt is given below, corresponding to a summer noon scenario with high solar elevation, moderate power output, increasing trends, and partly cloudy conditions:

\begin{Verbatim}
task=pv_power_forecasting | target=power |
output=per_station_multi_step_sequence |
forecast_steps=48 | step_minutes=15 |
context_short_steps=12 | context_long_steps=48 |
site_latitude_band=latitude_38_to_40 |
site_longitude_band=longitude_113_to_115 |
season=summer | time_of_day=noon | solar_elevation=high |
power_last_bin=medium | power_trend_short=increase |
power_trend_long=increase | power_variability_short=stable |
irradiance_last_bin=high | irradiance_trend_short=increase |
irradiance_trend_long=increase | irradiance_variability_short=variable |
temperature_last_bin=high | humidity_last_bin=medium |
wind_speed_last_bin=low | pressure_last_bin=medium |
cloud_indicator=likely_cloudy |
irradiance_power_coherence_short=positive |
irradiance_to_power_response_delay=delay_1_step |
forecast_observation_irradiance_gap=small
\end{Verbatim}

\bibliographystyle{elsarticle-num} 
\bibliography{Manuscript}

\end{document}